\documentclass[10pt,letterpaper]{article}
\usepackage{multirow}
\usepackage[final]{neurips_2020}
\usepackage{times}
\usepackage{epsfig}
\usepackage{graphicx}
\usepackage{amsmath}
\usepackage{amssymb}
\usepackage[ruled,vlined]{algorithm2e}
\usepackage{caption}
\usepackage{subcaption}
\usepackage{color,soul}

\usepackage[breaklinks=true,bookmarks=false]{hyperref}

\begin{document}
	
	\title{ Residual Error: a New Performance Measure for Adversarial Robustness}
	\author{
  \normalfont{Hossein Aboutalebi\textsuperscript{1}, Mohammad Javad Shafiee\textsuperscript{1,3}}\\
Michelle Karg\textsuperscript{2}, Christian Scharfenberger\textsuperscript{2}\\
 Alexander Wong\textsuperscript{1,3}\\
 \textsuperscript{1}Waterloo AI Institute, University of Waterloo, Waterloo, Ontario, Canada\\
  \textsuperscript{2}ADC Automotive Distance Control Systems GmbH, Continental, Germany \\
  $^3$DarwinAI Corp., Canada\\
  \textsuperscript{1}\{haboutal, mjshafiee, a28wong\}@uwaterloo.ca \\
  \textsuperscript{2}\{michelle.karg, christian.scharfenberger\}@continental-corporation.com
}

	\maketitle
	\begin{abstract}
		Despite the significant advances in deep learning over the past decade, a major challenge that limits the wide-spread adoption of deep learning has been their fragility to adversarial attacks.  This sensitivity to making erroneous predictions in the presence of adversarially perturbed data makes deep neural networks difficult to adopt for certain real-world, mission-critical applications.  While much of the research focus has revolved around adversarial example creation and adversarial hardening, the area of performance measures for assessing adversarial robustness is not well explored.  Motivated by this, this study presents the concept of residual error, a new performance measure for not only assessing the adversarial robustness of a deep neural network at the individual sample level, but also can be used to differentiate between adversarial and non-adversarial examples to facilitate for adversarial example detection.  Furthermore, we introduce a hybrid model for approximating the residual error in a tractable manner.  Experimental results using the case of image classification demonstrates the effectiveness and efficacy of the proposed residual error metric for assessing several well-known deep neural network architectures.  These results thus illustrate that the proposed measure could be a useful tool for not only assessing the robustness of deep neural networks used in mission-critical scenarios, but also in the design of adversarially robust models.
	\end{abstract}

\section{Introduction}

Deep learning models over the past few years have yield new records in many fields such as  different computer vision applications~\cite{krizhevsky2012imagenet, lecun2010convolutional, he2016deep, redmon2016you, redmon2018yolov3}, machine translation~\cite{wu2016google, vaswani2017attention},  and medicine~\cite{de2018clinically, cruz2017accurate}. Although these achievements bring new level of accuracy, recent research especially within computer vision applications have shown that deep neural networks are unable to detect some of the intuitive underlying concepts in datasets~\cite{szegedy2013intriguing, goodfellow2014explaining, moosavi2016deepfool}. These findings generally follow the idea that adding small perturbation $\epsilon$ to the input image causes the model to classify the image incorrectly. The perturbation to the input image is imperceptible to human eyes most of the time. This phenomenon was first discovered by szegedy {\it et al.} in their seminal paper~\cite{szegedy2013intriguing}. They observed that the state-of-the-art deep neural networks act poorly  with high confidence when an imperceptible non-random perturbation is added to the input image. The perturbed examples so-called \textit{"adversarial examples"} are generated by adding a targeted noise calculated based on the loss value and projected gradient. They also discovered that the adversarial examples are shared between different network architectures and training data. In other words, if a set of adversarial examples are generated for one specific network, it is possible that these adversarial examples will still be missclassified by another network with different architecture even if the new network is trained on different training data. 

Goodfellow {\it et al.}~\cite{goodfellow2014explaining} attributed this poor performance of deep neural networks on adversarial examples to their linear behavior in high-dimensional spaces. Using this observation, they proposed a fast method of generating adversarial examples, \textit{``Fast Gradient Sign Method''} (FGSM). 
Given the input $x$, the target $y$ and the objective function $l(\theta,x,y)$ that deep neural network with parameters $\theta$ tries to minimize, FGSM generates the following perturbation: 
\begin{align}
\label{fgsm}
\eta= \epsilon \cdot \text{sign}\Big(\nabla_x l(\theta,x,y)\Big)
\end{align}
where max-norm of $\eta$ is constrained to keep the perturbation imperceptible to human eyes. 

Moosavi-Dezfooli {\it et al.}~\cite{moosavi2016deepfool} proposed another method to generate adversarial examples which induces less  amount of noise on the input image compared to FGSM. They  achieved this result through an iterative process in which at each step the function is linearized around the input.

Finally, Madry {\it et al.}~\cite{madry2017towards} proposed a new algorithm of adversarial example generation which is able to produce adversarial examples that are harder to learn. The proposed method is based on solving a min-max optimization problem and it generalizes the prior first-order adversarial algorithms.

Broadly speaking, adversarial attacks can be divided into two categories of black-box attacks and white-box attacks. In the black-box attacks, the algorithm does not have access to the target model and its parameters~\cite{ilyas2018black,papernot2017practical} while, the white-box attackers have direct access to target model and as such they can generate more severe adversarial perturbations~ \cite{rakin2018parametric}. 

Parallel to introducing new adversarial perturbation algorithms, several approaches have been proposed to improve the robustness of deep neural network facing adversarial attacks as well. 
One of the well-know approach to mitigate the effect of adversarial attack is to take advantage of adversarial training~\cite{szegedy2013intriguing, kurakin2016adversarial} which has been shown to increase the robustness of a network by augmenting the training data with adversarial examples.
Nevertheless, building a robust deep neural network against adversarial attacks may cause a drop in natural accuracy. Su {\it et al.}~\cite{su2018robustness} studied this trade-off between adversarial robustness and natural
accuracy and found out that 
a higher performance in testing accuracy causes a reduction in robustness. Moreover, Zhang {\it et al.}~\cite{zhang2019theoretically} provided a theoretical analysis of the trade-off between accuracy and robustness in deep neural networks. In their work, they provide an upper bound on the gap between robust error and natural accuracy.

Here, we formulate the problem of deep neural networks facing  adversarial attacks within a different point of view.  Given the trained model, we argue that due to the existence of adversarial examples, the  test error is not a sufficient measure to indicate the accuracy of the trained model. As such, we propose a new measure that can be used besides the test error to estimate the model error for individual input so-called  residual error. Specifically our contributions in this paper are as follows:

\begin{itemize}
    \item Introduction of a new performance measure for adversarial robustness called residual error which provides a more precise measure of error at the individual sample level.
    \item Proposing a novel prediction model learning method for approximating the residual error.
    \item Designing a novel hybrid model which can estimate residual error in an accurate manner while being an order of magnitude faster to execute.
    \item Introduction of a novel strategy to harnessing residual error prediction model for detecting adversarial examples.
    \item Comprehensive experimental results to evaluate the efficacy of the proposed residual error measure and associated prediction models for the task of image classification on the CIFAR-10 and CIFAR-100 datasets.
\end{itemize}

The paper is organized as follows.  The underlying theory behind residual error along with the methodology for approximating the residual error by learning a hybrid prediction model is described in detail in Section 2. The experimental results are presented and discussed in detail in Section 3.  Finally, conclusions are drawn and future directions are discussed in Section 4.

\section{Methodology}
In this section, we will describe in detail the underlying theory behind the proposed residual error performance metric.  Furthermore, we will describe in detail how the residual error can be approximated using a prediction model, and further introduce a hybrid model for estimating residual error.

\subsection{Residual Error}

Let us assume the hypothesis $h \in \mathcal{H}$\footnote{$\mathcal{H}$ is the set of hypothesis} is a mapping function to model $\mathcal{X} \xrightarrow{} \mathcal{Y}$. As such, $h$ tries to estimate the target function  $f(\cdot)$ based on the training data $S \subset \mathcal{X}\times \mathcal{Y}$; therefore, the error of $h$ with loss function $l\big(h(x), f(x)\big)$ is defined as follows:
\begin{align}
\label{err1}
L_{D,f}(h)= \mathbb{E}_{x \sim D}
\Big[ l(h(x), f(x))\Big]
\end{align}
where $D$ is the distribution of domain space. For a regression problem with MSE error, we can rewrite \eqref{err1} as follows:
\begin{align}
\label{err2}
L_{D,f}(h)= \mathbb{E}_{x \sim D} \Big[ \big(h(x)- f(x)\big)^2\Big]
\end{align}
and the error can be calculated for a classification problem using the indicator function:
\begin{align}
\label{err3}
L_{D,f}(h)= \mathbb{E}_{x \sim D} \left[ \textbf{1}_{h(x) \neq f(x)}\right]
\end{align}
where $ \textbf{1}_{h(x) \neq f(x)}$ is equal to $1$ only when the predicted label for input $x$ by hypothesis $h$ is different from the target label $f(x)$.

However, the exact value of \eqref{err1} cannot be calculated and as such, a test set is used to estimate its value. Assuming $S'$ is our test set, the test error becomes:
\begin{align}\label{err4}
L_{S',f}(h)= \frac{1}{|S'|} \sum_{x \in S'} l(h(x), f(x))
\end{align}
which is the empirical error approximates  the final accuracy of the trained model in practice.

Considering the existence of adversarial examples, in this paper, we argue that the test error may not be the sufficient measure of empirical error of the model. In this regard, \eqref{err1} is the value of the expected error over the domain space $\mathcal{X}$ and \eqref{err1} does not provide any further information for individual input $x$. This  also holds for the test error as it is the empirical estimation of \eqref{err1}. For example, although the trained model behaves differently under adversarial attack, the test error does not provide an insight to differentiate between adversarial and non-adversarial example. As a result, we need to provide a precise estimate of the model error for an individual input $x$:
\begin{align} \label{err5}
R_h(x)=\mathbb{E}\Big[l\big(h(x),f(x)\big)\Big]
\end{align}
which \eqref{err5} is  called the residual error of the trained model $h$ for input $x$. 

The benefit of having residual error $R_h(x)$ to the test error $L_{S',f}(h)$ is that now we can decide based on the input if the model is able to make the correct prediction or not. While test error provides a general error estimation on the whole domain, residual error gives us an error measure on each individual input data from the domain.  A particular interpretation that we can have is that if $R_h(x)$ is a high value and the model has a low error on the test set, we can expect that the input $x$ might be an adversarial example. 
This way, we will be able to detect adversarial examples. We will cover this adversarial example detection feature further in the experiment section 3.1.1.

As such, the main remaining challenge here is that how we want to estimate the value of $R_h(x)$. We will cover this in the next section.

\subsection{Residual Error Prediction Model}

Estimating $R_h(x)$ is different from estimating the test error since the training set used for this estimation is only a subset which represents the domain partially. As such,  approximating $R_h(x)$ is highly desirable. Given the prediction model $h$, the residual error of trained model $h$ can be  approximated using prediction function $g(\cdot)$.
Here, we call $h$ the primary model and $g$ as the residual error prediction model. The residual error prediction model is  trained given the primary model has been already trained.

We now describe the dataset used to train the residual error prediction model. For each pair of $(x,y)\in S$ in our training set,  it is replaced  with $(x,r(x))$, where 
\begin{align}\label{r0}
r(x)=l\big(h(x),y\big).
\end{align}
As a result, the training dataset for the residual error prediction model is constructed as follow:
\begin{align}\label{r1}
S_{res}=\Big\{(x,r(x))~~|~~ (x,y) \in S ~~\wedge ~~ r(x)=l\big(h(x),y\big) \Big\}.
\end{align}

For a regression problem, the labels of training data model is represented by the MSE error and the residual error prediction model training is still a regression problem. On the other hand, the labels of  the residual error prediction model become $0$ or $1$ (i.e., the primary model classifies the sample correctly or not) and the training  of the residual error prediction model is a classification problem, and it is formulated as a binary classification problem. In this regard, although the primary model training might be done with a multi-class classification training data, the residual error prediction model training becomes a simpler task of binary classification. 

The loss function for residual error prediction model is the Cross-entropy loss for the classification problem, while the MSE loss can be used to  train the residual error prediction model for a regression problem. It is worth to note that other variation of loss functions with regularization can be used for training the residual error prediction model if the loss function can fit the learning problem of the residual error prediction model.

In our experiments, we have found that  using the whole training data to train the residual error prediction model may not be effective. Recent advances on the state-of-the-art deep neural networks which facilities training of these models with almost zero error in the training step reduces the number of useful training data to train the residual error prediction model. To address this issue,  we  use the validation set to train the residual error prediction model in our experimental setup. It turns out that although the validation set might be a fraction of the training data, the validation size is still sufficient to train the residual error prediction model as the problem of residual error prediction model is simpler than the primary model especially for the case of classification problems. Moreover, the validation set is augmented by adversarial examples to mitigate the problem of imbalance classes in training residual error prediction model (i.e., most of samples are classified correctly by the primary model and, therefore, there are more samples with class 1 rather than class 0). The details of the training of the residual error prediction model is described in Algorithm~\ref{alg:one}\footnote{$S_{valid}$ refers to validation set}.

\begin{algorithm}[!t]
	\KwData{$\Big\{\big(x,r(x)\big)| x \in D\Big\}$}
	\KwResult{$g^*$: trained residual error prediction model $g$ }
	\textbf{input}: validation set: $S_{valid}$\\ ~~~~~~~~~~~~primary model: $h$\\ ~~~~~~~~~~~~residual error prediction model :$ g $ \\ 
	~~~~~~~~~~~~loss function primary model: $ l $ \\ 	
	~~~~~~~~~~~~loss function residual prediction model: $l' $ \\ 	
	\textbf{begin}\\
	~~~~~$S$=[]\\
	~~~~~ \textbf{for} $(x,y)$ in $S_{valid}$:\\
	~~~~~~~~~~$r(x)=l\big(y,h(x)\big)$\\
	~~~~~~~~~~Insert $(x,r(x))$ into $S$\\
	~~~~~ $g^*$ = $\underset{g}{arg min} ~ \sum_{x \in S} l'\big(g,r(x)\big)$\\
	~~~~~ \textbf{return}~ $g^*$\\
	\textbf{end}
	
	\caption{Resedual Network Training}
	\label{alg:one}
\end{algorithm}

\subsection{Hybrid Residual Error Prediction Model}
In this section, we propose a hybrid architecture design which takes advantage of  deep neural network macroarchitecture to extract useful features from the input image while the decision tree macroarchitecture tries to discriminate the decision-making of the primary network in whether it classifies the input image correctly or not.     
Our observations have shown that using the proposed hybrid structure as the residual error prediction model can improve the performance of the proposed measure. The proposed hybrid model can use the primary model architecture to extract useful features followed by a decision tree for the classification purposes. As such, the output layer (classifier) of the primary module is substituted by a decision tree where the output of the last layer  in the primary model is fed into the decision tree. During the training of residual error prediction model, the primary model weights are frozen and the decision tree parameters are only  updated. 

\begin{figure*}
\centering
	\includegraphics[scale=0.6]{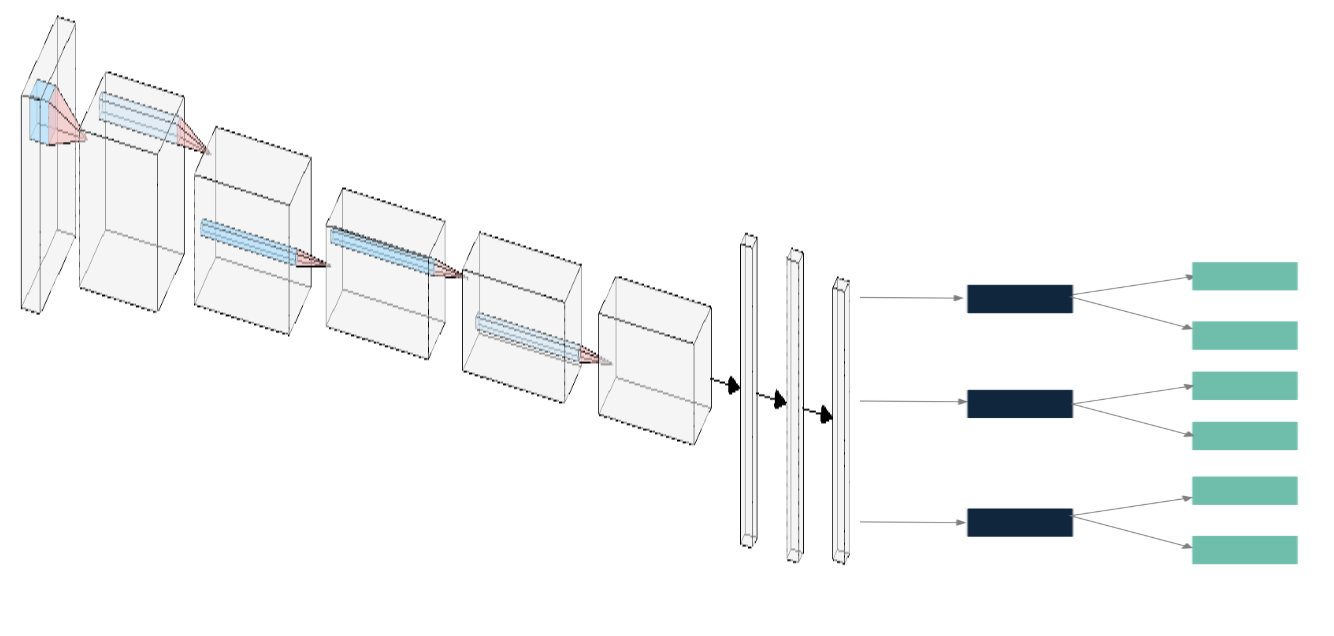}
	\caption{The architecture design of the proposed residual error prediction model. The first architectural component of the prediction model consists of a deep neural network macroarchitecture design, while the second component consists of a decision tree macroarchitecture to produce the final residual error prediction. }
	\label{fig:model}
\end{figure*}

This approach can benefit from the feature representation of a deep neural network while reducing the whole problem of training the residual error prediction model with only constructing a decision tree. Experimental results showed that the training of the proposed hybrid residual  model is an order of magnitude faster than training a deep network model from scratch and use as the residual error prediction model. In our experiments, we used XGBoost~\cite{chen2016xgboost} decision tree structure. The architecture of our model is depicted in Figure~\ref{fig:model}.

\section{Experimental Results}

\begin{figure*}
	\begin{tabular}{cc}
		\includegraphics[width=0.5\textwidth]{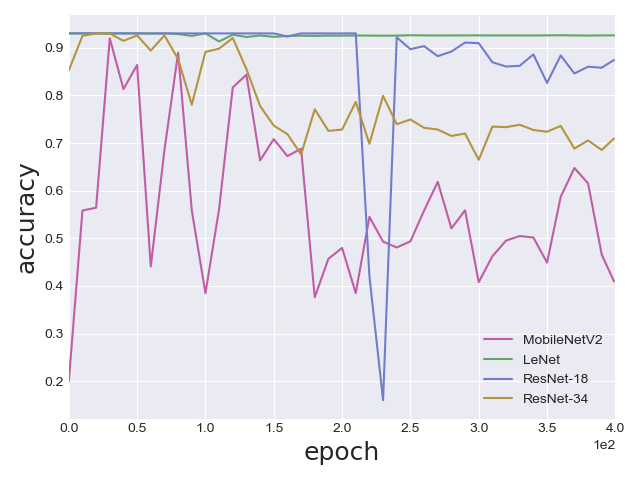}&
		\includegraphics[width=0.5\textwidth]{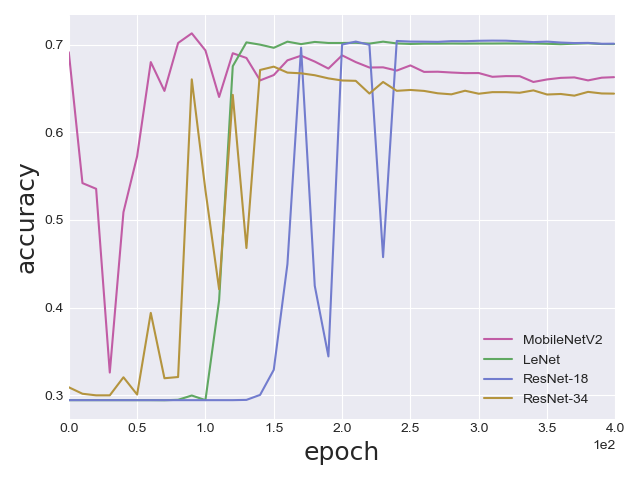}\\
		(a) Normal Dataset  & (b) Adversarial Dataset
\end{tabular}
	\caption{Residual error prediction model accuracy on CIFAR-10. The primary model is ResNet-18.}
	\label{fig:curve1}
\end{figure*}

\begin{figure*}
	\begin{tabular}{cc}
		\includegraphics[width=0.5\textwidth]{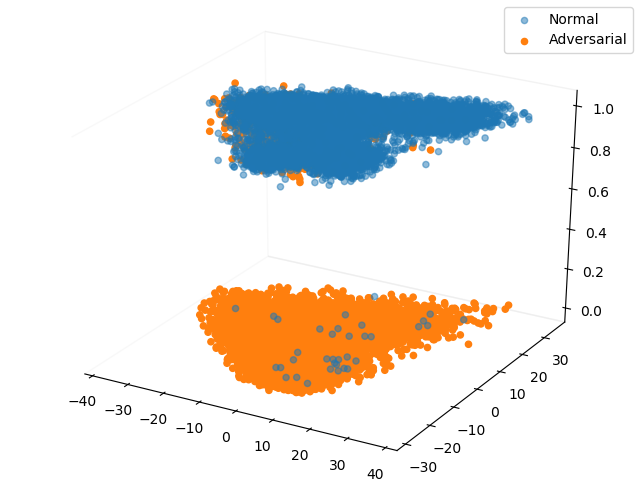}&
		\includegraphics[width=0.5\textwidth]{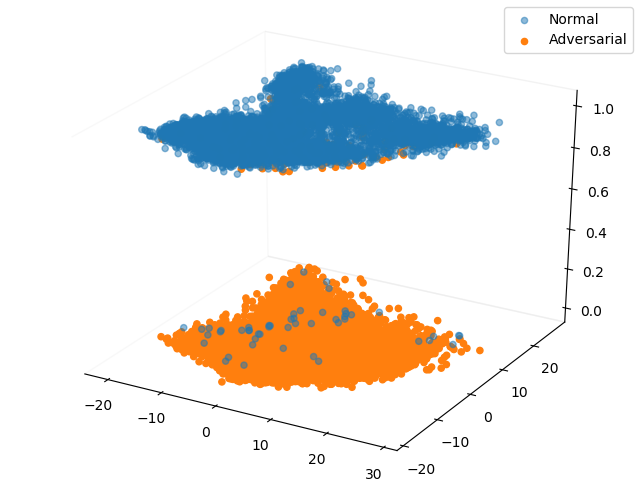}\\
		(a)   & (b) 
\end{tabular}
	\caption{The clustering of dataset with residual error prediction model into normal and adverserial clusters. Residual error prediction model in both figures is LeNet and primary model in figure (a) is MobileNetV2 and figure (b) is ResNet-18 and the dataset is CIFAR-10. The LeNet is able to detect Normal datapoint from adversarial datapoint.}
	\label{fig:cluster}
\end{figure*}

The proposed method is evaluated based on different neural network architectures and  with two datasets of CIFAR-10 and CIFAR-100. Different neural network architectures including ResNet-18, ResNet-34 \cite{he2016deep}, LeNet and MobileNetV2 \cite{sandler2018mobilenetv2} are used to measure the effectiveness of the proposed method. The proposed hybrid model is compared with other approaches as well. To examine the performance of the proposed residual method the input samples are perturbed by FGSM with $\epsilon=8$ adversarial attacks.
\subsection{CIFAR-10}
Table~\ref{tab:cifar10} shows the experimental results of evaluating the proposed residual error prediction model  on different deep neural network architecture on CIFAR-10 dataset. To better evaluate the effectiveness of the proposed method, the performance of the residual error prediction model is assessed  on two different situations, i) the normal dataset where the examined samples are clean and without any adversarial perturbation, ii) the adversarial dataset which is the samples are perturbed by FGSM adversarial attacks. The goal here is to determine what is performance of the residual error prediction model in detecting whether the primary network is classifying the samples correctly or not. The training data for the  residual error prediction model is created based on  the residual validation set explain in~\eqref{r1}. As mentioned in Section 3, in order to increase the size of training data for training the residual error prediction model, the residual error prediction models are trained by both clean and  adversarial examples. 

A grid search algorithm was done to choose the best hyper-parameters for the residual error prediction models. As seen in Table~\ref{tab:cifar10}, although the primary models perform with relatively low accuracy on adversarial examples, the residual error prediction models could identify whether the primary models is performing correctly or not with a very higher performance. The experimental results show that the proposed hybrid model which is the composite of a primary model and XGBoost outperforms  other deep networks (except for LeNet as the primary model) in adversarial dataset experiments. The hybrid model provided the best average performance compared to other residual error prediction models in these experiments. 

One of the counter-intuitive observation in our experiments is that although we were expecting better performance from bigger network architectures like ResNet-34, their performance is not as good as smaller ones   like LeNet. We were first suspecting that one of the reason for this observation was because of the hyperparamters. However, after running extensive experiments, we found that this is not the case. Indeed, due to the size of the residual dataset $S_{res}$ for residual error prediction model training, larger model tends to get very high accuracy on $S_{res}$ and start overfitted to the noise. We found that early stopping is an effective way to prevent such an unwanted behavior. 

\begin{figure*}
\begin{tabular}{cc}
		\includegraphics[width=0.5\textwidth]{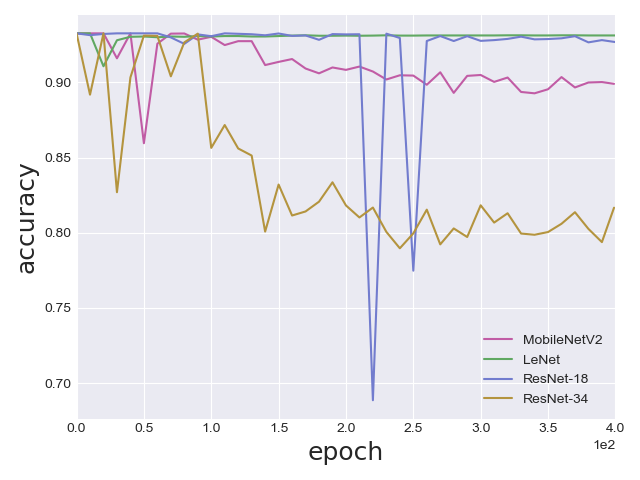}&
		\includegraphics[width=0.5\textwidth]{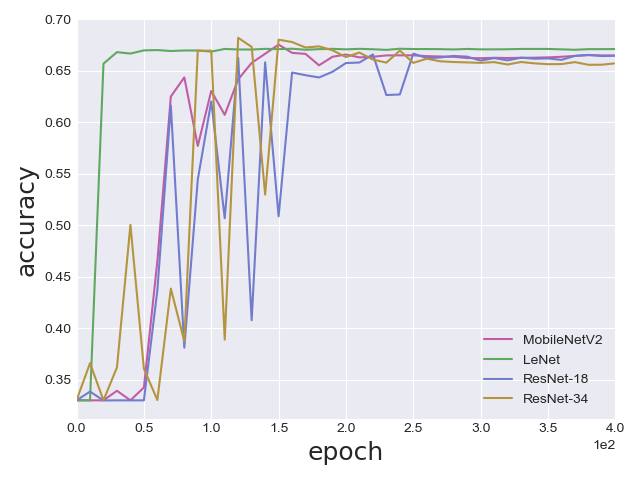}\\
		(a) Normal Dataset  & (b) Adversarial Dataset
		
\end{tabular}
\caption{Residual error prediction model accuracy on CIFAR-10. The primary model is ResNet-34.}
\label{fig:curve2}
\end{figure*}

Figure~\ref{fig:curve1} and Figure~\ref{fig:curve2} demonstrate  the learning curves of residual error prediction models through different training epochs. As seen in Figure~~\ref{fig:curve1} (a) and Figure~\ref{fig:curve2} (a), ResNet-34 and MobileNetV2 have compromised their accuracy on normal dataset to get higher accuracy on adversarial samples. These two Figures clearly show the need for early stopping in ResNet-34 and MobileNetV2.

\subsection{CIFAR-100}
To better evaluate the effectiveness of the proposed residual error prediction model, the same experiments are conducted for the CIFAR-100 as well. Different network architectures are used for both primary and residual error prediction models. Table~\ref{tab:cifar100} shows the accuracy of different network architectures based on normal and adversarial datasets. Due to the very low accuracy of LeNet model (i.e., because of its low capacity) on  CIFAR-100, it is excluded as a primary model in this experiment. Unlike what we observed in CIFAR-10, we can see that ResNet-34 has a higher accuracy for adversarial examples compared to other residual error prediction models (except for ResNet-18 where it is the second best model). On the other hand, Hybrid model has the best accuracy on the normal dataset.

\subsection{Adversarial Example Detection}
Finally, Figure~\ref{fig:cluster} shows the effectiveness of the proposed method in  discriminating adversarial examples from non-adversarial examples. The primary model used in this experiment (a) is MobileNetV2 and ResNet-18 in experiment (b) in Figure~\ref{fig:cluster};  LeNet architecture was used as the residual error prediction model architecture. To better visualize the samples the image tensors are projected into a 2D space using PCA algorithm where the z-axis is the output of  LeNet in predicting whether the sample is normal or adversarial sample. As seen, the residual model outputs 1 almost on all non-adversarial examples and it outputs 0 on adversarial examples. The proposed method can almost perfectly cluster adversarial and non-adversarial examples as shown in Figure~\ref{fig:cluster}. As such, the proposed residual model can be used to identify whether the input sample fed into the primary model is perturbed for the purpose of adversarial attack or not. 

\begin{table*}[ht] \label{tab:cifar10}
	\hspace{-1.5cm}
	\begin{tabular}{|c c c || c c c c c|} 
		\hline
		
		\multirow{ 2}{*}{Primary model} & \multirow{ 2}{*}{Dataset} &\multirow{ 2}{8em} {Primary  Accuracy} & \multicolumn{5}{|c|}{Residual error prediction model}  \\
		
		&&& ResNet-18  & ResNet-34  & LeNet &  MobileNetV2 & Hybrid\\ [0.5ex] 
		\hline\hline
		\multirow{ 2}{*}{\bf ResNet-18}  & Normal & 0.9304 & 0.8747 & \textbf{0.9303}  & 0.9262 & 0.8844& 0.8761 \\ 
		&   Adversarial & 0.2999 & 0.7012  & 0.7067& 0.7007 & 0.7051  & \textbf{0.7503}\\ 
		\hline
		\multirow{ 2}{*}{\bf ResNet-34}  & Normal & 0.9327  & 0.9086 &0.8321 & \textbf{0.9314} & 0.8991 & 0.8882 \\ 
		&   Adversarial & 0.3302 & 0.6609 & 0.6804& 0.671 &0.6649  & \textbf{0.7471} \\ 
		\hline
		\multirow{ 2}{*}{\bf LeNet} & Normal & 0.746 & 0.6946 &  0.3971 & \textbf{0.7248} & 0.3687 & 0.6303 \\ 
		&   Adversarial & 0.0353 & 0.3055 &0.9483& \textbf{0.9545} & 0.9454& 0.8921 \\ 
		\hline
		\multirow{ 2}{*}{\bf  MobileNetV2}  & Normal &  0.9201  & 0.8414 &  0.8402& \textbf{0.9176} & 0.8730 & 0.8856 \\ 
		&   Adversarial & 0.3248 &  0.6848  & 0.6649 & 0.6752& 0.6923  &\textbf{0.7852}\\ 
		\hline
	\end{tabular}
	\caption{Results of residual error prediction model accuracy on CIFAR-10. Here adversarial refers to FGSM attack with $\epsilon=8$, and normal dataset is the dataset without any adversarial examples.}
	\label{tab:cifar10}
\end{table*}
\begin{table*}[ht]
	\hspace{-1.5cm}
	\begin{tabular}{|c c c || c c c c c|} 
		\hline
		
		\multirow{ 2}{*}{Primary model} & \multirow{ 2}{*}{Dataset} &\multirow{ 2}{8em} {Primary  Accuracy} & \multicolumn{5}{|c|}{Residual error prediction model}\\
		
		&&& ResNet-18  & ResNet-34  & LeNet &  MobileNetV2 & Hybrid\\ [0.5ex] 
		\hline\hline
		\multirow{ 2}{*}{\bf ResNet-18}  & Normal & 0.7454 & 0.7454  & 0.7454  & 0.7436 & 0.6994 & \textbf{0.7604} \\ 
		&   Adversarial & 0.156 & 0.7183  & 0.9131& 0.9124 & \textbf{0.9145}  & 0.8755\\ 
		\hline
		\multirow{ 2}{*}{\bf ResNet-34}  & Normal & 0.7579  & 0.7106 &0.7482 & 0.7486 & 0.4441 & \textbf{0.7763}  \\ 
		&   Adversarial &0.1233 & 0.876 & \textbf{0.8752} & 0.8736 &0.8063  & 0.8458 \\ 
		\hline
		\multirow{ 2}{*}{\bf MobileNetV2}  & Normal &  0.7067  & 0.7064 &  0.6992& 0.7057 & 0.6989 & \textbf{0.7092} \\ 
		&   Adversarial & 0.0781 &  0.9219  & \textbf{0.923}  & 0.9222& 0.9111  &0.8922\\ 
		\hline
	\end{tabular}
	\caption{Results of residual error prediction model accuracy on CIFAR-100. Here adversarial refers to FGSM attack with $\epsilon=8$, and normal dataset is the dataset without any adversarial examples.}
	\label{tab:cifar100}
\end{table*}

\section{Conclusion}

In this work, we presented the notion of residual error, a new performance metric for not only assessing adversarial robustness at the individual sample level but also differentiating between adversarial and non-adversarial examples, thus facilitating for adversarial example detection. A hybrid prediction model comprised of deep neural network  and decision tree macroarchitectures is  to improve the performance of the residual model to mitigate the lack of training data and improve the effeciency of the model.  The proposed performance metric is especially useful with the existence of adversarial attacks as it can provide a confidence bound on the performance of the trained deep neural network for each input. Experimental results  showed the performance of the hybrid residual error prediction model on several different image classification networks. Building better hybrid residual error prediction models with higher accuracy is an interesting direction for future research.  Furthermore, there are many applications of residual error  beyond the context of adversarial robustness assessment, as it can also be harnessed as a safety measure in other domains of machine learning, which would be interesting to explore as a future direction.

{\small
\bibliographystyle{plainnat}
\bibliography{egbib}

\begin{thebibliography}{21}
\providecommand{\natexlab}[1]{#1}
\providecommand{\url}[1]{\texttt{#1}}
\expandafter\ifx\csname urlstyle\endcsname\relax
  \providecommand{\doi}[1]{doi: #1}\else
  \providecommand{\doi}{doi: \begingroup \urlstyle{rm}\Url}\fi

\bibitem[Chen and Guestrin(2016)]{chen2016xgboost}
Tianqi Chen and Carlos Guestrin.
\newblock Xgboost: A scalable tree boosting system.
\newblock In \emph{Proceedings of the 22nd acm sigkdd international conference
  on knowledge discovery and data mining}, pages 785--794, 2016.

\bibitem[Cruz-Roa et~al.(2017)Cruz-Roa, Gilmore, Basavanhally, Feldman,
  Ganesan, Shih, Tomaszewski, Gonz{\'a}lez, and Madabhushi]{cruz2017accurate}
Angel Cruz-Roa, Hannah Gilmore, Ajay Basavanhally, Michael Feldman, Shridar
  Ganesan, Natalie~NC Shih, John Tomaszewski, Fabio~A Gonz{\'a}lez, and Anant
  Madabhushi.
\newblock Accurate and reproducible invasive breast cancer detection in
  whole-slide images: A deep learning approach for quantifying tumor extent.
\newblock \emph{Scientific reports}, 7:\penalty0 46450, 2017.

\bibitem[De~Fauw et~al.(2018)De~Fauw, Ledsam, Romera-Paredes, Nikolov, Tomasev,
  Blackwell, Askham, Glorot, O’Donoghue, Visentin, et~al.]{de2018clinically}
Jeffrey De~Fauw, Joseph~R Ledsam, Bernardino Romera-Paredes, Stanislav Nikolov,
  Nenad Tomasev, Sam Blackwell, Harry Askham, Xavier Glorot, Brendan
  O’Donoghue, Daniel Visentin, et~al.
\newblock Clinically applicable deep learning for diagnosis and referral in
  retinal disease.
\newblock \emph{Nature medicine}, 24\penalty0 (9):\penalty0 1342--1350, 2018.

\bibitem[Goodfellow et~al.(2014)Goodfellow, Shlens, and
  Szegedy]{goodfellow2014explaining}
Ian~J Goodfellow, Jonathon Shlens, and Christian Szegedy.
\newblock Explaining and harnessing adversarial examples.
\newblock \emph{arXiv preprint arXiv:1412.6572}, 2014.

\bibitem[He et~al.(2016)He, Zhang, Ren, and Sun]{he2016deep}
Kaiming He, Xiangyu Zhang, Shaoqing Ren, and Jian Sun.
\newblock Deep residual learning for image recognition.
\newblock In \emph{Proceedings of the IEEE conference on computer vision and
  pattern recognition}, pages 770--778, 2016.

\bibitem[Ilyas et~al.(2018)Ilyas, Engstrom, Athalye, and Lin]{ilyas2018black}
Andrew Ilyas, Logan Engstrom, Anish Athalye, and Jessy Lin.
\newblock Black-box adversarial attacks with limited queries and information.
\newblock \emph{arXiv preprint arXiv:1804.08598}, 2018.

\bibitem[Krizhevsky et~al.(2012)Krizhevsky, Sutskever, and
  Hinton]{krizhevsky2012imagenet}
Alex Krizhevsky, Ilya Sutskever, and Geoffrey~E Hinton.
\newblock Imagenet classification with deep convolutional neural networks.
\newblock In \emph{Advances in neural information processing systems}, pages
  1097--1105, 2012.

\bibitem[Kurakin et~al.(2016)Kurakin, Goodfellow, and
  Bengio]{kurakin2016adversarial}
Alexey Kurakin, Ian Goodfellow, and Samy Bengio.
\newblock Adversarial machine learning at scale.
\newblock \emph{arXiv preprint arXiv:1611.01236}, 2016.

\bibitem[LeCun et~al.(2010)LeCun, Kavukcuoglu, and
  Farabet]{lecun2010convolutional}
Yann LeCun, Koray Kavukcuoglu, and Cl{\'e}ment Farabet.
\newblock Convolutional networks and applications in vision.
\newblock In \emph{Proceedings of 2010 IEEE international symposium on circuits
  and systems}, pages 253--256. IEEE, 2010.

\bibitem[Madry et~al.(2017)Madry, Makelov, Schmidt, Tsipras, and
  Vladu]{madry2017towards}
Aleksander Madry, Aleksandar Makelov, Ludwig Schmidt, Dimitris Tsipras, and
  Adrian Vladu.
\newblock Towards deep learning models resistant to adversarial attacks.
\newblock \emph{arXiv preprint arXiv:1706.06083}, 2017.

\bibitem[Moosavi-Dezfooli et~al.(2016)Moosavi-Dezfooli, Fawzi, and
  Frossard]{moosavi2016deepfool}
Seyed-Mohsen Moosavi-Dezfooli, Alhussein Fawzi, and Pascal Frossard.
\newblock Deepfool: a simple and accurate method to fool deep neural networks.
\newblock In \emph{Proceedings of the IEEE conference on computer vision and
  pattern recognition}, pages 2574--2582, 2016.

\bibitem[Papernot et~al.(2017)Papernot, McDaniel, Goodfellow, Jha, Celik, and
  Swami]{papernot2017practical}
Nicolas Papernot, Patrick McDaniel, Ian Goodfellow, Somesh Jha, Z~Berkay Celik,
  and Ananthram Swami.
\newblock Practical black-box attacks against machine learning.
\newblock In \emph{Proceedings of the 2017 ACM on Asia conference on computer
  and communications security}, pages 506--519, 2017.

\bibitem[Rakin et~al.(2018)Rakin, He, and Fan]{rakin2018parametric}
Adnan~Siraj Rakin, Zhezhi He, and Deliang Fan.
\newblock Parametric noise injection: Trainable randomness to improve deep
  neural network robustness against adversarial attack.
\newblock \emph{arXiv preprint arXiv:1811.09310}, 2018.

\bibitem[Redmon and Farhadi(2018)]{redmon2018yolov3}
Joseph Redmon and Ali Farhadi.
\newblock Yolov3: An incremental improvement.
\newblock \emph{arXiv preprint arXiv:1804.02767}, 2018.

\bibitem[Redmon et~al.(2016)Redmon, Divvala, Girshick, and
  Farhadi]{redmon2016you}
Joseph Redmon, Santosh Divvala, Ross Girshick, and Ali Farhadi.
\newblock You only look once: Unified, real-time object detection.
\newblock In \emph{Proceedings of the IEEE conference on computer vision and
  pattern recognition}, pages 779--788, 2016.

\bibitem[Sandler et~al.(2018)Sandler, Howard, Zhu, Zhmoginov, and
  Chen]{sandler2018mobilenetv2}
Mark Sandler, Andrew Howard, Menglong Zhu, Andrey Zhmoginov, and Liang-Chieh
  Chen.
\newblock Mobilenetv2: Inverted residuals and linear bottlenecks.
\newblock In \emph{Proceedings of the IEEE conference on computer vision and
  pattern recognition}, pages 4510--4520, 2018.

\bibitem[Su et~al.(2018)Su, Zhang, Chen, Yi, Chen, and Gao]{su2018robustness}
Dong Su, Huan Zhang, Hongge Chen, Jinfeng Yi, Pin-Yu Chen, and Yupeng Gao.
\newblock Is robustness the cost of accuracy?--a comprehensive study on the
  robustness of 18 deep image classification models.
\newblock In \emph{Proceedings of the European Conference on Computer Vision
  (ECCV)}, pages 631--648, 2018.

\bibitem[Szegedy et~al.(2013)Szegedy, Zaremba, Sutskever, Bruna, Erhan,
  Goodfellow, and Fergus]{szegedy2013intriguing}
Christian Szegedy, Wojciech Zaremba, Ilya Sutskever, Joan Bruna, Dumitru Erhan,
  Ian Goodfellow, and Rob Fergus.
\newblock Intriguing properties of neural networks.
\newblock \emph{arXiv preprint arXiv:1312.6199}, 2013.

\bibitem[Vaswani et~al.(2017)Vaswani, Shazeer, Parmar, Uszkoreit, Jones, Gomez,
  Kaiser, and Polosukhin]{vaswani2017attention}
Ashish Vaswani, Noam Shazeer, Niki Parmar, Jakob Uszkoreit, Llion Jones,
  Aidan~N Gomez, {\L}ukasz Kaiser, and Illia Polosukhin.
\newblock Attention is all you need.
\newblock In \emph{Advances in neural information processing systems}, pages
  5998--6008, 2017.

\bibitem[Wu et~al.(2016)Wu, Schuster, Chen, Le, Norouzi, Macherey, Krikun, Cao,
  Gao, Macherey, et~al.]{wu2016google}
Yonghui Wu, Mike Schuster, Zhifeng Chen, Quoc~V Le, Mohammad Norouzi, Wolfgang
  Macherey, Maxim Krikun, Yuan Cao, Qin Gao, Klaus Macherey, et~al.
\newblock Google's neural machine translation system: Bridging the gap between
  human and machine translation.
\newblock \emph{arXiv preprint arXiv:1609.08144}, 2016.

\bibitem[Zhang et~al.(2019)Zhang, Yu, Jiao, Xing, Ghaoui, and
  Jordan]{zhang2019theoretically}
Hongyang Zhang, Yaodong Yu, Jiantao Jiao, Eric~P Xing, Laurent~El Ghaoui, and
  Michael~I Jordan.
\newblock Theoretically principled trade-off between robustness and accuracy.
\newblock \emph{arXiv preprint arXiv:1901.08573}, 2019.

\end{thebibliography}
}

\end{document}